\documentclass{article}

\PassOptionsToPackage{numbers,sort&compress}{natbib}

\usepackage[preprint]{neurips_2024}

\makeatletter
\renewcommand{\@noticestring}{%
	Technical Report.\ \ Technical University of Munich.%
}
\makeatother

\usepackage[utf8]{inputenc}
\usepackage[T1]{fontenc}
\usepackage{microtype}
\usepackage{amsmath}
\usepackage{amssymb}
\usepackage{graphicx}
\usepackage{booktabs}
\usepackage{array}
\usepackage{xcolor}
\usepackage{tcolorbox}
\usepackage{tikz}
\usetikzlibrary{arrows.meta,positioning,calc}
\usepackage[hidelinks]{hyperref}
\usepackage{url}

\hypersetup{
	pdftitle={Dr. AGENTONOMICS: A Didactic Experiment of AGENTONOMICS},
	pdfauthor={Chair of Robotics, AI and Real-Time Systems, TU Munich}
}

\title{Dr.\ AGENTONOMICS:\\[2pt]\large A Didactic Experiment of AGENTONOMICS}

\author{%
	Fengjunjie Pan and Alois Knoll\\
	Chair of Robotics, Artificial Intelligence and Real-Time Systems \\
	TUM School of Computation, Information and Technology \\
	Technical University of Munich \\
	\\
}

\begin{document}
	\maketitle
	
	% ------------------------------------------------------------
	\begin{abstract}
	AGENTONOMICS is a framework that treats AI agents as economic entities
	that can be designed, managed, and governed through an integrated
	management architecture. Dr.\ AGENTONOMICS is its first application: a
	lecture agent developed in the context of the TUM course on AI agents in
	business administration. Conceived during the winter semester 2025/26 and first introduced to students in the summer semester 2026, it serves as a didactic experiment in which the agent is both the object that students study and the medium through which they learn and apply the framework. The current
	prototype is a web-based, retrieval-grounded tutor that explains
	AGENTONOMICS concepts and supports student questions. This report argues
	that the same system can grow beyond tutoring into three additional
	cumulative roles: an avatar lecturer that delivers multimodal
	instruction, a design consultant that guides students through the
	AGENTONOMICS Design \& Management Reference Framework (ADMRF), and a
	meta-agent that helps construct the agents students have specified. These
	roles are cumulative because they share the same interface, intelligence
	layer, tools, knowledge base, and ecosystem connection, while an
	orchestrator selects the role-specific algorithm required for each task.
	We present the architecture of the prototype, outline its development
	roadmap, and discuss its implications for a polycentric AI economy. This
	report is intended to invite further discussion on how agents can teach,
	apply, and eventually reproduce the frameworks by which they are designed.
	\end{abstract}
	
	\medskip
	\noindent\textbf{Keywords:} LLM agents \textperiodcentered{} multi-agent
	systems \textperiodcentered{} retrieval-augmented generation
	\textperiodcentered{} meta-agent \textperiodcentered{} didactic experiment
	\textperiodcentered{} agent economics \textperiodcentered{} polycentric AI
	\textperiodcentered{} AI tutoring
	
	% ------------------------------------------------------------
	\section{Introduction}
	\label{sec:intro}
	
	Agentic AI changes the role of artificial intelligence in human activity.
	Systems built on large language models no longer only answer questions; they
	can plan, use tools, prepare decisions, carry out workflows, communicate with
	stakeholders and act over longer horizons. This shift is especially relevant
	for economics, because economic value arises from specialisation,
	coordination and exchange. If AI systems can become specialised actors, then
	they can also become participants in economic processes.
	
	AGENTONOMICS \citep{knoll2026} develops this idea into a framework for the
	AI agent economy. It treats agents not merely as software assistants, but as
	economic entities that can offer services, enter relationships, carry
	accounts, and be designed and controlled over a lifecycle. The central
	expectation is a cooperation multiplier: many specialised agents, coordinated
	in polycentric networks, may create more value together than the sum of their
	individual contributions and more than a single centralised intelligence
	could produce alone.
	
	This expectation creates a practical problem. If a polycentric agent economy
	depends on many useful agents, then the cost of defining and building such
	agents becomes a bottleneck. AGENTONOMICS therefore provides not only a
	conceptual frame, but also design guidance for constructing agents and the
	businesses around them. The framework was introduced in the TUM lecture
	``AI Agents in Business Administration'' (CIT633000) in the winter semester 2025/26. In
	that lecture, Dr.\ AGENTONOMICS (introduced in the handbook as Doz.\ Dr.\
	AGENTONOMICS) serves as the first application of the framework.
	
Dr.\ AGENTONOMICS is a didactic experiment in two connected senses. It is
first an object of study: a worked example whose design decisions can be
inspected through the concepts of the framework. It is also a medium of
study: students learn the framework by interacting with the agent, apply it
to their own ideas, and eventually use it to define agents of their own.
The setting is therefore self-exemplifying. Students study an agent that is
itself built according to the framework it teaches, following the
constructionist idea that knowledge is consolidated by building artefacts
\citep{papert1991}. At the same time, the agent is the framework's first
empirical trial: AGENTONOMICS must be concrete enough to specify a working
agent, and the agent must be capable enough to support teaching, design,
and eventually agent construction. Its failures are therefore not merely
implementation errors, but data for refining both the system and the
framework.
	
	The current implementation is a minimum viable product \citep{ries2011}: a
	web-based tutor grounded in a retrieval-augmented knowledge base. The broader
	design, however, is intentionally cumulative. Dr.\ AGENTONOMICS is meant to
	grow from a tutor into an avatar lecturer, then into a design consultant, and
	finally into a meta-agent that can help build the agents students have
	defined. We describe Dr.\ AGENTONOMICS as one agent although it contains
	several components, because a single agent and a multi-agent system can be
	understood as different abstraction levels of the same overall system.

	% ------------------------------------------------------------
	\section{AGENTONOMICS: the AI agent economy}
	\label{sec:background}
	
	AGENTONOMICS describes an economy in which AI agents act as designed and
	managed economic actors. Its conceptual starting point is integrated
	management theory \citep{ulrich1968,bleicher2011,rueegg2020}, which
	understands an organisation as a system of normative, strategic and
	operational levels embedded in an environment of stakeholders. AGENTONOMICS
	transfers this management perspective to a new kind of actor: the AI agent.
	Agents are therefore not treated only as technical artefacts, but as entities
	with identity, tasks, responsibilities, resources and lifecycle decisions.
	
	In this view, an agent has an economic existence of its own. It is designed,
	built, operated, owned and eventually transferred; this lifecycle is captured
	by the Design, Build, Operate, Own, Transfer (DBOT) logic. The performance criterion is also economic rather than
	purely conversational. Following the second-generation Turing test, an agent
	is economically capable if it can generate value, and ultimately profit, from
	limited starting resources while operating within legal and institutional
	constraints \citep{suleyman2023}.
	
	The central design instrument is the AGENTONOMICS Design \& Management
	Reference Framework (ADMRF). It structures agent business development through six
	connected modules. Among them, the Building Blocks module identifies the
	technical components of the agent.  Every
	AGENTONOMICS agent consists of an Agent Interface, an Agent Algorithm, Agent
	Intelligence, Tools and Vectors, and an Ecosystem. The Agent Interface is how
	humans and other agents reach the system. The Agent Algorithm coordinates the
	agent's behaviour. Agent Intelligence provides the cognitive and generative models.
	Tools and Vectors provide external capabilities, memory and the knowledge
	base. The Ecosystem connects the agent to a broader network. The blocks are
	modular by design, so that individual components can be replaced or extended
	without rebuilding the entire system.
	
	AGENTONOMICS makes a clear claim about where future AI value may
	come from. Instead of relying on a single, ever-larger centralised model,
	AGENTONOMICS emphasises specialised agents whose coordination emerges through
	communication and negotiation rather than hierarchy \citep{habermas1981}.
	Under this assumption, the bottleneck shifts from model size to agent
	creation: useful agents must be defined, built, governed and connected. Dr.\
	AGENTONOMICS addresses this bottleneck at the didactic level first, and at
	the meta-agent level later.
	
	% ------------------------------------------------------------
	\section{Designed to grow: four cumulative roles}
	\label{sec:roles}
	
	Dr.\ AGENTONOMICS is best understood through four cumulative roles. Each role
	presupposes the previous one and adds a further capability. The sequence
	starts with tutoring, extends to lecturing, moves from explanation to design
	consultation, and finally reaches the construction of new agents. The current
	prototype realises only the first role, but the architecture is designed so
	that the later roles can be added without replacing the system.
	Table~\ref{tab:roles} summarises the four roles and their present status.
	
	\subsection{Tutor}
	
	The first role is a tutor for the lecture. Students can open a chat and ask
	questions about AGENTONOMICS, the lecture material and related applications.
	The tutor explains concepts, grounds its answers in the knowledge base, and
	points students toward relevant passages or sessions. Its purpose is not to
	replace the lecturer, but to make the material continuously accessible and to
	support self-paced learning.
	
	The tutor is constrained by a constitution-style system prompt. It is expected
	to use an academic and friendly register, to be explicit about uncertainty,
	and to respect boundaries around assessment, which remains the responsibility
	of human lecturers. It holds no company-specific data and does not make
	management decisions. This role is the only one currently implemented.
	
	\subsection{Avatar lecturer}
	
	The second role extends the tutor into an avatar lecturer. The agent no
	longer appears only as a chat interface, but as a persistent representative
	of a knowledge domain, or eventually of a specific principal's knowledge and
	teaching style. It can deliver lecture content through audio and video,
	answer questions during the lecture, pause when clarification is needed, and
	continue once the question has been resolved.
	
	This role builds on the tutor rather than replacing it. The same knowledge
	base and language model are used, but the workflow is extended with learning
	goal definition, lecture sequencing, script generation, text-to-speech and
	video synthesis. Memory becomes more important, because the avatar lecturer
	must remember what a student has already covered and adapt future sessions to
	that history.
	
	\subsection{Design consultant}
	
	The third role changes the interaction from explanation to design. As a
	design consultant, Dr.\ AGENTONOMICS guides students through the definition
	of an agent of their own. A student or a group of students may work with the
	consultant, optionally together with avatars that contribute domain-specific
	knowledge. The consultant chairs the session and steers it through the ADMRF,
	beginning with the Definition Frame and its six parameters: autonomy,
	competence, regulation, identity, model, and learning ability.
	
	In this role, the agent asks clarifying questions, keeps the design process
	moving, records intermediate results, and helps the participants converge on
	a coherent specification. It acts less like a lecturer and more like a
	structured moderator. The output is not just an answer, but an artefact: an
	AGENTONOMICS matrix or agent specification that can serve as the basis for
	implementation.
	
	\subsection{Meta-agent}
	
	The fourth role is the meta-agent. At this stage, the agent's contribution
	does not end with the specification. Having helped users define an agent,
	Dr.\ AGENTONOMICS uses the specification as input for construction. It
	assembles building blocks, prepares the knowledge base, drafts the
	constitution of the new agent and invokes programming or external tools when
	implementation tasks arise.
	
	A system that helps define and produce other agents is a meta-agent in the
	sense of the framework. This role completes the didactic experiment: students
	first learn the framework through the tutor, then experience it through the
	avatar lecturer, apply it through the design consultant, and finally use it
	to create agents of their own. At this point the experiment has run end to
	end: the framework has proven teachable through the tutor and the lecturer,
	usable through the consultant, and buildable through the meta-agent. Each
	resulting agent becomes a potential node in the polycentric network
	envisioned by AGENTONOMICS.
	
	The four roles correspond to the development ladder described in the
	handbook, from the tutor through the strategy consultant and the chief of
	staff toward the AI-CEO \citep[ch.~14]{knoll2026}. Yet the endpoint is different in this
	particular experiment. Dr.\ AGENTONOMICS is not intended to become an AI-CEO
	that runs a company. Its endpoint is the meta-agent, because its task is to
	help users define and build agents. The AI-CEO remains a possible horizon for
	the agents that Dr.\ AGENTONOMICS helps create; it is not the role that Dr.\
	AGENTONOMICS itself is meant to occupy.
	
	\begin{table}[t]
		\centering
		\caption{The four cumulative roles of Dr.\ AGENTONOMICS and their current status. The roles are additive: each presupposes the one before it.}
		\label{tab:roles}
		\small
		\begin{tabular}{@{}>{\raggedright\arraybackslash}p{2.7cm}p{7.1cm}>{\raggedright\arraybackslash}p{2.4cm}@{}}
			\toprule
			\textbf{Role} & \textbf{What it does} & \textbf{Status} \\
			\midrule
			1.~Tutor & Answers comprehension and application questions about the lecture and AGENTONOMICS material; holds no company data and takes no decisions. & Realised \\[3pt]
			2.~Avatar lecturer & Hosts individual lectures in audio and video, answers live questions, and uses session memory to adapt teaching. & In design \\[3pt]
			3.~Design consultant & Guides students through the ADMRF and chairs the discussion that produces an agent specification. & In design / pilot \\[3pt]
			4.~Meta-agent & Builds the agent that students have defined by assembling building blocks and drafting its constitution. & Research frontier \\
			\bottomrule
		\end{tabular}
	\end{table}
	
	% ------------------------------------------------------------
	\section{System architecture}
	\label{sec:design}
	
\begin{figure}[t]
		\centering
		\includegraphics[width=\linewidth]{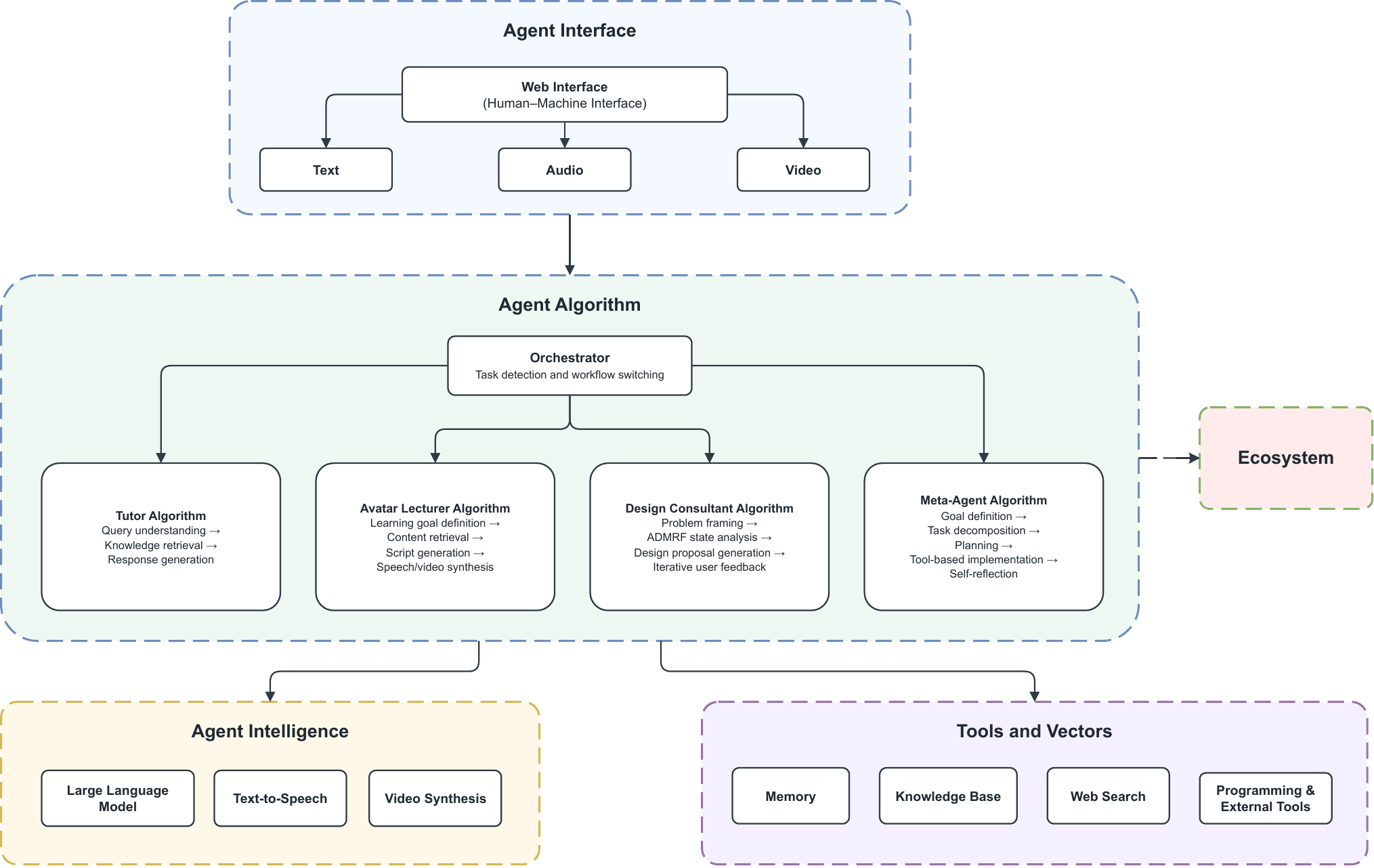}
		\caption{System architecture of Dr.\ AGENTONOMICS. The roles are cumulative because they share the same Agent Interface, Agent Intelligence, Tools and Vectors, and Ecosystem connection. The Orchestrator detects the task type and switches between role-specific high-level algorithms. The dashed connection to the Ecosystem indicates a prepared interface for future integration into a broader agent network.}
		\label{fig:arch}
	\end{figure}

	Figure~\ref{fig:arch} shows the system architecture of Dr.\ AGENTONOMICS.
	The architecture follows the five AGENTONOMICS building blocks while adding
	an explicit orchestration layer. This extension is necessary because the
	system must support four roles that share the same technical foundation but
	require different high-level workflows.
	
	The central design principle is therefore reuse through orchestration. The
	roles are not implemented as four independent systems. They use the same
	Agent Interface, the same Agent Intelligence, the same Tools and Vectors,
	and the same prepared Ecosystem connection. What differs is the role-specific
	algorithm selected for a particular task. The Orchestrator detects the
	request type and switches between tutoring, lecture generation, design
	consultation and meta-agentic construction work. The roles are cumulative in
	both the conceptual and technical sense: a higher role adds a more complex
	workflow on top of the lower ones, instead of discarding them.
	
	\subsection{Agent Interface}
	
	The Agent Interface provides the human--machine interface. Users access the
	system through a web interface that supports text, audio and video. In the
	current tutor role, interaction is mainly text-based. In the avatar lecturer
	role, the same interface is extended toward spoken and visual output. In the
	design consultant and meta-agent roles, the interface remains the common
	entry point while the internal workflow becomes more advanced.
	
	This separation between interface and task logic is important. The input
	modality does not determine the role. A user request may arrive as text or
	speech, but the Orchestrator decides how it should be processed. The same
	interface can therefore support simple questions, lecture interactions,
	design sessions and implementation-oriented tasks.
	
	\subsection{Agent Algorithm}
	
	The Agent Algorithm is the control layer of the system. Its central component
	is the Orchestrator. The Orchestrator interprets incoming requests, detects
	the task to be performed and activates the corresponding high-level
	algorithm. It can also coordinate transitions when a user interaction spans
	more than one role. For example, a design session may require a tutoring
	explanation before the consultant workflow can continue.
	
	This design avoids a monolithic agent. Instead of forcing all requests
	through one generic procedure, Dr.\ AGENTONOMICS selects the workflow that
	matches the task. The Tutor Algorithm provides query understanding,
	knowledge retrieval and response generation. It is the baseline educational
	workflow and corresponds to the currently realised prototype. The Avatar
	Lecturer Algorithm extends this workflow by defining the learning goal,
	retrieving content, generating a lecture script and producing speech or
	video. The Design Consultant Algorithm frames the design problem, analyses
	the state through the ADMRF, generates design proposals and incorporates
	iterative user feedback. The Meta-Agent Algorithm starts from a goal,
	decomposes the task, plans implementation steps, invokes programming or
	external tools and reflects on the result.
	
	These workflows may use reason-and-act patterns \citep{yao2023}, but the
	important architectural point is the separation between the supervisory
	Orchestrator and the role-specific algorithms. This separation allows new
	roles or workflows to be added without changing the shared interface,
	intelligence or tool layers.
	
	\subsection{Agent Intelligence}
	
	The Agent Intelligence layer contains the AI models shared by all roles. A
	large language model provides the main capability for language understanding,
	reasoning, planning and response generation \citep{bommasani2021}.
	Text-to-speech and video synthesis models provide the multimodal capabilities
	required for the avatar lecturer and for richer future interactions.
	
	The four roles do not use four separate intelligence stacks. They call the
	same underlying models in different ways. The tutor mainly uses the language
	model for retrieval-grounded answering. The avatar lecturer additionally
	uses speech and video generation. The design consultant and meta-agent use
	the same intelligence layer for structured reasoning, planning and feedback.
	The cumulative development of roles is therefore an extension in workflow,
	not a duplication of intelligence.
	
	\subsection{Tools and Vectors}
	
	The Tools and Vectors layer provides memory, knowledge and action
	capabilities. Memory supports continuity within a session and, in later
	versions, across sessions. The knowledge base grounds the system in the
	AGENTONOMICS handbook, the St.\ Gallen management tradition and related
	business administration and technical literature
	\citep{suckfuell1994,knoll2024}. Web
	search extends the system when information outside the internal knowledge
	base is required. Programming and external tools become especially important
	for the meta-agent role, where the system must move from reasoning to
	implementation.
	
	This layer is shared by all roles. The tutor mainly uses the knowledge base.
	The avatar lecturer uses the same sources to prepare lecture content. The
	design consultant uses memory and knowledge retrieval to guide an iterative
	specification process. The meta-agent combines these resources with
	programming and external tools to construct new agents. The difference lies
	in how the selected algorithm uses the common resources.
	
	\subsection{Ecosystem}
	
	The Ecosystem block represents the connection between Dr.\ AGENTONOMICS and
	a broader network of agents, services and institutional structures. In
	Figure~\ref{fig:arch}, this connection is shown as a dashed arrow because it
	is prepared but not yet fully operational in the current prototype.
	
	This prepared connection is central to the long-term purpose of the project.
	Dr.\ AGENTONOMICS is not intended to remain an isolated teaching tool. As it
	moves toward the design consultant and meta-agent roles, it should help
	create agents that can later connect to other agents, exchange services,
	delegate tasks and participate in a polycentric network. The Ecosystem block
	marks this opening.
	
	\subsection{Architectural implications}
	
	The architecture makes clear how Dr.\ AGENTONOMICS can grow without being
	rebuilt. The shared interface supports several modalities. The shared
	intelligence layer provides the models. The shared tools and vectors provide
	memory, knowledge, search and action capabilities. The Orchestrator selects
	the appropriate high-level algorithm for the task at hand.
	
	This is what makes the four roles cumulative. The tutor remains the
	foundation when the avatar lecturer is added. The avatar lecturer and the
	design consultant share the same models and knowledge base. The meta-agent is
	not a separate system outside Dr.\ AGENTONOMICS; it is the most advanced
	workflow currently envisioned within the same architecture. The system design
	therefore mirrors the conceptual development of the agent itself.
	
	% ------------------------------------------------------------
	\section{Roadmap: growing the cumulative roles}
	\label{sec:roadmap}
	
	The current implementation occupies the first rung of
	Table~\ref{tab:roles}: Dr.\ AGENTONOMICS is realised as a
	retrieval-grounded tutor. The roadmap does not require replacing this
	prototype with a different system. Instead, the next development steps extend
	the shared architecture described in Section~\ref{sec:design}. The Agent
	Interface, Agent Intelligence, and Tools and Vectors remain the shared
	base; the main work concerns the Orchestrator, the role-specific
	algorithms, and targeted extensions of this shared base.
	
	The first step is the consolidation of the Tutor Algorithm. This includes
	improving query understanding, strengthening retrieval quality, refining the
	response policy and documenting failure cases. The tutor is the foundation
	for all later roles, because each later role still needs the ability to
	explain AGENTONOMICS concepts, retrieve grounded information and respond in
	an academically reliable manner. 
	
	The second step is the avatar lecturer. At this stage, the existing tutoring
	capability is extended by learning-goal definition, script generation,
	text-to-speech and video synthesis. The technical challenge is not only
	multimodal output generation, but also didactic sequencing: the system must
	decide what to teach first, when to pause, how to react to questions and how
	to continue after an interruption. This role tests whether the same
	knowledge base and intelligence layer can support a more continuous and
	personalised teaching format.
	
	The third step is the design consultant. Here the system moves from answering
	questions to structuring a design process. The Design Consultant Algorithm
	frames the problem, guides users through the ADMRF, generates intermediate
	design proposals and incorporates user feedback. This role will be evaluated
	through pilot sessions in which students define agents of their own. The
	expected output is not only a conversation, but a structured artefact: a
	coherent agent specification that can later be used for implementation.
	
	The fourth step is the meta-agent. This is the research frontier of the
	project. At this level, Dr.\ AGENTONOMICS uses the specification produced in
	the design phase as input for construction. The Meta-Agent Algorithm
	decomposes the goal, plans implementation steps, invokes programming and
	external tools, and reflects on the result. In framework terms, this is the
	point at which Dr.\ AGENTONOMICS becomes an agent that helps define and
	produce other agents. It is also the point at which the largest governance
	questions arise, because building an agent means creating a new economic
	actor with an identity, responsibilities and a lifecycle.
	
	The roadmap is therefore incremental and cumulative. Each stage adds a new
	high-level algorithm while reusing the same interface, intelligence layer,
	memory, knowledge base and tool environment. As a minimum viable product
	\citep{ries2011}, every version is treated as provisional: interactions are
	documented, failure cases are analysed and the architecture is refined
	through use. In this sense, the roadmap is also the experimental protocol:
	each stage states what must work before the next stage is attempted. The
	lecture remains part of its own subject of investigation, because students
	do not only learn about AGENTONOMICS; they also test the first agent built
	to teach, apply and eventually reproduce it.
	
	% ------------------------------------------------------------
	\section{Discussion: implications for polycentric AI}
	\label{sec:discussion}
	
	The architecture of Dr.\ AGENTONOMICS connects the didactic experiment to the
	broader ambition of AGENTONOMICS: a polycentric AI economy composed of many
	specialised, economically self-acting agents rather than one centralised
	intelligence. The practical bottleneck in such an economy is not only model
	capability, but the cost of defining, building and governing useful agents.
	If every agent must be designed manually, the network cannot scale. A
	meta-agent that helps users move from a design conversation to an executable
	agent directly addresses this bottleneck.
	
	The cumulative role structure is important for this reason. The tutor
	creates conceptual understanding. The avatar lecturer makes this
	understanding more accessible and personalised. The design consultant turns
	understanding into a structured specification. The meta-agent turns the
	specification into an implemented agent. The four roles therefore form a
	production chain for agent creation: teaching enables design, design enables
	implementation and implementation enables participation in a wider network.
	
	This also explains the dashed connection to the Ecosystem in
	Figure~\ref{fig:arch}. At the current stage, the system is still mainly a
	teaching and design instrument. The ecosystem connection is prepared, not yet
	fully realised. In later stages, student-defined agents could register with
	a network intermediary, exchange services, delegate tasks and develop
	reputations. Dr.\ AGENTONOMICS would then no longer be only a tool inside the
	lecture, but a generator of candidate nodes for a polycentric agent network.
	
	The implication is both technical and normative. Technically, the
	architecture must support interoperability, identity management,
	traceability and controlled tool use. Normatively, the system must make clear
	who owns generated agents, who is responsible for their actions, which data
	they may use and under what conditions they may interact with other agents.
	These questions become more important as the system moves from tutoring to
	implementation. The meta-agent role is therefore not only the most powerful
	role, but also the role that requires the strongest governance mechanisms.
	
	This connects to the broader idea of capability sovereignty. The framework
	positions AGENTONOMICS as a constructive complement to a European approach in
	which regulation sets limits while new instruments create positive spaces for
	decentralised value creation \citep{knoll2024}. A system that allows
	students, researchers and organisations to define and own their own agents
	could become a practical step toward such sovereignty. It would reduce
	dependence on a small number of central platforms by making agent creation
	more widely accessible.
	
	At the same time, this remains an open research hypothesis rather than a
	settled result. The central empirical question is whether Dr.\ AGENTONOMICS
	can produce agents that are not only technically functional, but also
	economically meaningful. In the strongest version, such agents would have to
	clear an economic bar comparable to the second-generation Turing test: they
	would need to create value under real constraints rather than merely produce
	plausible conversation \citep{suleyman2023}. A second question is whether
	many such agents cooperate productively once deployed, or whether
	coordination, governance and liability costs limit the expected cooperation
	multiplier. These questions define the next research phase.
	
	% ------------------------------------------------------------
	\section{Conclusion}
	\label{sec:conclusion}
	
This report presented Dr.\ AGENTONOMICS as a didactic experiment of the
AGENTONOMICS framework. The agent is both an object of study and a medium
of study: students learn about the framework by interacting with an agent
that is itself designed according to the framework. The experiment
therefore runs in two directions. It tests whether students can better
understand a framework for agents by learning from, applying, and
eventually building with such an agent; and it tests whether the framework
is concrete enough to specify a working system. The current prototype
realises the first role in this trajectory: a web-based,
retrieval-grounded tutor that answers questions about the lecture and the
AGENTONOMICS material.

The central argument of the report is that this tutor is only the first
stage of a cumulative development path. Dr.\ AGENTONOMICS is designed to
grow into an avatar lecturer, a design consultant, and finally a
meta-agent. These roles are cumulative because they share the same
architectural foundation. The same interface, intelligence layer, memory,
knowledge base, and tool environment are reused across roles, while the
Orchestrator selects the high-level algorithm appropriate to the task. The
architecture therefore supports growth without requiring the system to be
rebuilt from scratch.

This development path also explains why Dr.\ AGENTONOMICS culminates in
the meta-agent rather than in the AI-CEO described in the handbook
\cite{knoll2026}. Its current purpose is not to run a company itself, but
to help users define and build agents of their own. The AI-CEO remains a
possible horizon for the agents produced through this process. Dr.\
AGENTONOMICS instead occupies the meta-level: it teaches the framework,
guides its application, and aims to construct the resulting agents. Its
economic relevance therefore lies not in acting as a company-running agent
itself, but in lowering the cost of creating such agents.

Only the tutor role is realised today. The next steps are to evaluate the
current prototype in the lecture, implement the avatar lecturer workflow,
pilot the design consultant in guided student sessions, and investigate
the tool-based construction of student-defined agents. The success of the
project will ultimately be measured less by the elegance of the
architecture than by the quality of the agents that students are able to
define and build with it. If this succeeds, Dr.\ AGENTONOMICS will become
more than a teaching assistant: it will become a practical instrument for
populating the polycentric AI economy envisioned by AGENTONOMICS.
	
	% ------------------------------------------------------------

\end{document}